\RequirePackage[T1]{fontenc}
\documentclass[letterpaper,10pt,conference]{ieeeconf}

\IEEEoverridecommandlockouts
\usepackage{microtype}
\usepackage{amsmath,amssymb,booktabs,graphicx,url,balance}

\usepackage{hyperref}
\hypersetup{
    hidelinks,
    pdftitle={VCTP: Vehicle-Conditioned Terrain Planning for Off-Road Navigation},
    pdfauthor={Akshay Naik, Ramavarapu S. Sreenivas, Dustin Nottage, Ahmet Soylemezoglu},
    pdfsubject={Preprint submitted to the 2027 IEEE International Conference on Robotics and Automation (ICRA)}
}

\title{\LARGE\bfseries
VCTP: Vehicle-Conditioned Terrain Planning\\
for Off-Road Navigation
}

\author{%
Akshay Naik$^{1}$, Ramavarapu S.~Sreenivas$^{2}$,
Dustin Nottage$^{3}$, and Ahmet Soylemezoglu$^{3}$%
\thanks{This work has been submitted to the IEEE for possible publication.
Copyright may be transferred without notice, after which this version may
no longer be accessible.}%
\thanks{$^{1,2}$University of Illinois Urbana-Champaign,
Urbana, IL, USA:
$^{1}$Department of Electrical and Computer Engineering;
$^{2}$Department of Industrial and Enterprise Systems Engineering.}%
\thanks{$^{3}$U.S. Army Engineer Research and Development Center,
Construction Engineering Research Laboratory,
Champaign, IL, USA.}%
\thanks{E-mail (A.~Naik): \texttt{akshayn3@illinois.edu}.}%
}

\begin{document}

\maketitle
\thispagestyle{empty}
\pagestyle{empty}

\begin{abstract}
A vehicle's heading affects both the surfaces beneath its tires and its pitch and roll.
We present Vehicle-Conditioned Terrain Planning (VCTP), which retains these relationships by evaluating shared elevation and surface-ID layers at eight headings.
Body geometry constrains admissibility, while loaded wheel contacts determine modeled surface cost and predicted pitch and roll.
Established D* Lite and vehicle-state search use these evaluations to plan routes with forward and reverse motion.
When observations change, VCTP recomputes every affected body or contact query.
In fully observed two-track simulations, sampling at wheel contacts rather than at the vehicle center lowers modeled surface cost by 39.3\% while shortening the route.
In offline planning on RGator recordings, VCTP also reduces modeled costs over identified surfaces and observed support relative to distance-focused planning, although incomplete coverage leaves full-route rankings unresolved.
Selective updates match full recomputation in all 474 comparisons using recorded map changes.
These results identify when wheel-contact placement and vehicle heading affect route choice.
\end{abstract}
\section{Introduction}
\label{sec:introduction}

A terrain map alone does not determine where a vehicle can travel.
Our \emph{terrain map} combines a 2.5D elevation layer describing environment
geometry with a \emph{surface map} identifying pavement, grass, gravel and
other surface IDs. We interpret these observations using vehicle dimensions, motion constraints and tire properties. A passage may fit one vehicle but not another;
an open hillside may impose excessive roll. Vehicle parameters remain fixed
during a run.

We describe this interaction through three coupled cost characteristics.
\emph{Environment geometry} presents obstacles, structures and height
discontinuities. \emph{Vehicle fit and support} depend on the footprint,
turning radius and wheel placement, which determine maneuverability and
predicted pitch and roll. \emph{Tire--surface interaction} relates the
supporting surface to tire dimensions and loading. They overlap because
clearance, support attitude and tire placement depend on the same pose.

Hard safety limits exclude motions; finite costs rank the remaining choices.
A longer route may reduce unfavorable surface contact or roll. Crossing a
slope differs from ascending it, and a centerline on pavement can still place
a tire on grass. Center-only surface sampling and heading-averaged attitude
costs discard information that changing a weight cannot restore.

Observations arrive gradually: an obstruction beyond a bend may invalidate
a route. Missing geometry and surface IDs must remain distinct from observed
ground. We ask: \emph{how can fixed vehicle parameters translate
an updating terrain map into consistent route costs and feasibility decisions?}

Semantic mapping, risk-aware planning and body- or capability-aware
traversability provide important foundations
\cite{maturana2017semantic,fan2021step,guo2026rbtrg,capezzuto2026cat}.
We introduce \emph{Vehicle-Conditioned Terrain Planning} (VCTP) around one
premise: terrain cost depends on a vehicle's pose and contacts, not a map
cell alone. Its contributions are:
\begin{enumerate}
 \item A vehicle-conditioned representation retaining heading-dependent body
 clearance, support attitude and loaded-contact surface preference when
 querying shared terrain layers.
 \item Selective evaluation that invalidates and recomputes every vehicle
 query whose body or contact neighborhood overlaps changed observations,
 preserving consistency with full evaluation on the tested updates.
 \item Matched-input comparisons isolating when discarded heading/contact
 relationships change routes: fully observed simulation establishes the
 mechanisms; offline planning on field-recorded RGator data tests their
 consequences with imperfect sensing.
\end{enumerate}
Existing D* Lite and vehicle-state search use this representation for
forward/reverse planning; the search algorithms themselves are not new.

\begin{figure*}[t]
 \centering
 \includegraphics[width=\textwidth]{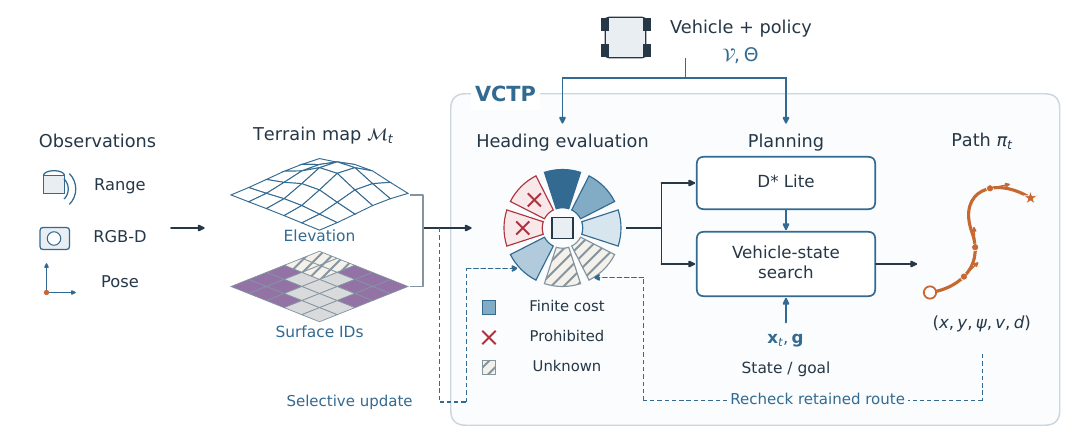}
 \caption{VCTP: shared terrain evidence, vehicle-specific evaluation and
 route generation. Each of eight headings determines body and contact placement.
 Dashed links mark selective updates and retained-route checks. Map and path
 sketches are schematic.}
 \label{fig:system_method}
\end{figure*}

\section{Related Work}
\label{sec:related}

\noindent\textit{Terrain representation.}
Probabilistic elevation mapping propagates sensor and localization
uncertainty~\cite{fankhauser2018probabilistic}. Semantic 2.5D mapping and
TNS combine image and range observations for off-road navigation
\cite{maturana2017semantic,guan2022tns}. TerrainNet learns geometric and
semantic features, while TravSUITE combines geometric--semantic mapping
with self-supervised costs and online risk adaptation
\cite{meng2023terrainnet,triest2026travsuite}. VCTP builds on this separation
of observations from traversal decisions. Its focus is where a vehicle
samples the evidence: a center label need not describe every tire, and a
local gradient need not describe support across a wheelbase.

\noindent\textit{Vehicle relationships with geometry.}
Footprint-aware planning evaluates orientation on rough ground
\cite{wermelinger2016navigation}; learned traversability can also condition
on robot properties and physical capabilities
\cite{eder2024robotdependent,capezzuto2026cat}. SPARTA models approach angle,
and Motion-aware Traversability conditions on velocity
\cite{dong2025sparta,zhao2026mat}. RB-TRG evaluates an oriented body along
motion and turning transitions, including support variation, lateral
inclination, interference and untrusted regions~\cite{guo2026rbtrg}.
These methods establish the value of vehicle-aware geometry. VCTP builds on
that foundation by evaluating surface IDs beneath loaded contacts together
with heading-dependent support. We adapt RB-TRG's body cost to the same
search and supply the same contact surface term. This comparison isolates
cost formulation within a shared planner.

\noindent\textit{Tire--surface costs.}
Bekker relates soil deformation to contact dimensions and loading
\cite{bekker1969terrain}; CAMIS derives directional costs from slope and
terramechanical effects~\cite{sanchez2022camis}. Experience-based methods
instead learn vehicle responses, including robot-dependent class costs and
proprioceptive or self-supervised traversal costs
\cite{eder2023traversability,castro2022feel,triest2025velociraptor}.
A preprint combines Bekker-inspired costs, wheel-contact attitude,
curvature-feasible planning and incremental repair~\cite{naik2025hybrid}.
Here we evaluate predicted attitude and surface preference at rotated,
load-weighted contacts within common body constraints. Matched-input
experiments test which retained relationships change route choice.

\noindent\textit{Incremental planning.}
D* Lite reuses prior search after graph changes~\cite{koenig2002dstar}; state
lattices and Hybrid A* encode feasible vehicle motion
\cite{pivtoraiko2009lattice,dolgov2008hybrida}. Systems such as STEP and TNS
already combine evolving maps with constrained planning
\cite{fan2021step,guan2022tns}. VCTP's update requirement follows from its
vehicle queries: changing one observation can affect neighboring poses
whose bodies or tires overlap it. Building on existing incremental search,
we test whether updating only these affected queries preserves the costs
and repair decisions obtained by full recomputation.

\section{Methodology}
\label{sec:method}

\subsection{Shared terrain evidence and vehicle queries}

Figure~\ref{fig:system_method} shows how observations become a vehicle-specific
route. Registered range data populate elevation $Z_t(c)$; image predictions
provide surface ID $S_t(c)$ and confidence $\gamma_t(c)\in[0,1]$ in the same
map frame. Updates populate a limited region of the persistent map $\mathcal M_t$.
Validity and obstruction evidence accompany ground support. This 2.5D model
excludes stacked surfaces and clearance beneath overhangs.

Vehicle parameters $\mathcal V$ and policy $\Theta$ define the query at
each of eight headings, $\psi_h=2\pi h/8$. Rotating a contact's body offset
$\mathbf p_j$ places it at $\mathbf x_j=c+R(\psi_h)\mathbf p_j$ on the map.
We estimate vehicle attitude by fitting a plane to bilinear height samples
$z_k=Z_t(c+R(\psi_h)\mathbf p_k)$ at body and contact patches.
With $\mathbf u_k=(p_{k,x},p_{k,y},1)^\top$, the fit is
\begin{equation}
 \mathbf q^*=\arg\min_{\mathbf q}\sum_k\omega_k
 [z_k-\mathbf u_k^\top\mathbf q]^2.
 \label{eq:plane}
\end{equation}
The four-wheel implementation uses 29 samples, weighted toward contacts.
For $\mathbf q=(a,b,z_0)^\top$, predicted pitch is $\alpha=\arctan(a)$ and
roll is $\beta=\arctan(b/\sqrt{1+a^2})$. Maximum fit residual $\rho$ tests
support consistency; height departure $\delta$ tests body interference.
An oriented body rectangle expanded by safe clearance $d_{\rm safe}$ tests
mapped obstructions. Observed violations of the declared pitch, roll,
residual, departure or clearance allowances exclude the cell--heading pair.
Unknown support has a separate flag. Search may provisionally enter these
cells, but they do not count as observed safe ground. The tests assess
quasi-static geometric compatibility.

\subsection{Directional policy cost}

A finite multiplier sets the cost per unit distance for each admissible
cell and heading:
\begin{equation}
 \begin{aligned}
 K(c,h)&=1+w_aF(c,h)+w_sT(c,h)+w_uU(c,h),\\
 F&=\left[\operatorname{clip}\!\left(
 \max\left\{\frac{|\alpha|}{\alpha_{\rm safe}},
             \frac{|\beta|}{\beta_{\rm safe}}\right\},0,1\right)\right]^2.
 \end{aligned}
 \label{eq:cost}
\end{equation}
The unit term prices distance. $F$ penalizes how much of the allowed pitch
or roll a pose uses, even below the safety angles. Inadmissible motion has
infinite cost. The three coupled relationships therefore enter through
both hard constraints and finite preferences.

For wheel radius $R_j$, width $b_j$, load $F_{z,j}$ and surface coefficients
$(k_{c,r},k_{\phi,r},n_r)$, the Bekker-inspired approximation
\cite{bekker1969terrain,naik2025hybrid} gives
\begin{equation}
 \begin{aligned}
 z_{jr}&=\left[\frac{F_{z,j}}
 {(k_{c,r}/b_j+k_{\phi,r})b_j\sqrt{2R_j}}\right]^{1/(n_r+1/2)},\\
 c_{jr}&=\operatorname{clip}(z_{jr}/R_j,0,1),\qquad
 \eta_j=F_{z,j}/\textstyle\sum_l F_{z,l},\\
 T&=\sum_j\eta_j\gamma(\mathbf x_j)c_{j,S_t(\mathbf x_j)},\qquad
 U=\sum_j\eta_j[1-\gamma(\mathbf x_j)].
 \end{aligned}
 \label{eq:surface}
\end{equation}
Here $c_{jr}$ is a normalized sinkage proxy used as a vehicle-parameterized
surface preference, not a validated prediction of realized sinkage or
traction. $\eta_j$ is the wheel's share of total load. Loads are quasi-static;
dynamic transfer is not modeled.
Literature-initialized coefficients use compatible units, and the normalization
stays fixed across methods and newly observed surfaces. Missing labels
contribute to $U$, not a low-cost ID. When support cannot be fitted, the
compiler sets $F=T=0$ and $U=1$ while retaining the unknown flag. Surface coefficients
stay fixed, but rotating the contacts can change the sampled surfaces and $T$.

\subsection{Motion and selective updates}

Planning has two stages. Grid-level D* Lite provides guidance
$\Pi_g$, pricing each edge by its length times the mean endpoint multiplier
and selecting among admissible forward/reverse headings. Vehicle-state
search then retains turning history and travel direction $d$ in states
$(x,y,\psi,d)$. Its route objective is
\begin{equation}
 \begin{split}
 J(\pi)={}&\sum_{e\in\pi}m(d_e)\int_0^{L_e}K(c_e(s),h(\psi_e(s)))\,ds\\
          &+L_gN_g+w_g\sum_{e\in\pi}d(c_e^{\rm end},\Pi_g).
 \end{split}
 \label{eq:path}
\end{equation}
Here $m$ is 1 forward and 1.25 reverse. Each gear change adds $L_g=2$\,m;
$w_g=0.25$ weights endpoint distance from grid guidance. Reverse changes
travel direction while preserving the physical contact layout. All methods
use the same search and these same terms. The discrete search does not
guarantee a globally optimal continuous trajectory.

Each Ackermann primitive joins straight tangents to a fixed-radius
$45^\circ$ arc and ends exactly at a lattice pose. Within a fixed $5\times5$
endpoint neighborhood, we choose the connection with the shortest added
tangents. Integration uses 10\,cm spacing on straights and 5\,cm on arcs.
A shared conservative swept-rectangle check rejects obstructed primitives,
including between samples, and search resumes within its original budget.
Only checked candidates are returned. These primitives bound curvature
without guaranteeing continuous steering rate.

With curvature $\kappa_i$ and length $\ell_i$, segment endpoint speeds obey
\begin{equation}
 \begin{aligned}
 v_i,v_{i+1}&\leq\min\{v_{\rm nom},\sqrt{a_{\rm safe}/|\kappa_i|}\},\\
 v_{i+1}^2-v_i^2&\leq2a_+\ell_i,\quad
 v_i^2-v_{i+1}^2\leq2a_-\ell_i.
 \end{aligned}
 \label{eq:speed}
\end{equation}
Forward and backward passes apply the acceleration and braking bounds;
the curvature cap is absent on straight segments.
Start, goal and both sides of a gear change have zero reference speed.

When a map cell changes, we recompute queries whose body or support/contact
neighborhood overlaps it and reuse the rest. Every validity change triggers
an update; finite-value changes use a $10^{-4}$ tolerance. An inadmissible
route is withdrawn before repair. A feasible route is retained despite
soft-cost changes, and failed repair issues no continuation. Obstruction
stamping remains global even when support evaluation is selective. We test
equality on recorded changes; sub-tolerance perturbations are outside that test.

\subsection{Evaluation criteria and compared methods}
\label{sec:evaluation}

Success is assessed by goal attainment, available safety checks and family
cost, alongside the required travel. Samples have distance weights $\ell_i$, with
$L=\sum_i\ell_i$. Elevation coverage $C_Z$ measures the route fraction with
observed body and support heights, excluding inferred starting support.
Surface coverage $C_S$ measures the fraction of tire travel with accepted
IDs, weighted by wheel load. Attitude cost $\bar F$ averages $F$ over
observed support. Surface cost $\bar T_{\rm known}$ averages $c_{j,S}$ over
identified contacts using weights $\ell_i\eta_j$; accepted IDs receive no
further confidence discount. Unknown portions remain unscored.

We also assess how much of the allowed turning demand a route uses:
$u_i=\max(R_{\rm turn}|\kappa_i|,v_{i,\rm peak}^2|\kappa_i|/a_{\rm safe})$,
using arc curvature and peak segment speed. Its route average is
$\bar G=L^{-1}\sum_i\ell_i\operatorname{clip}(u_i,0,1)^2$.
An assessment-only combined policy score is
\begin{equation}
 \begin{aligned}
 Q_\Theta&=\frac{w_a\bar F+w_s\bar T_{\rm known}+w_u\bar U+w_t\bar G}
 {w_a+w_s+w_u+w_t},\\
 \bar U&=L^{-1}\sum_i\ell_i\sum_j\eta_j(1-\gamma_{ij}).
 \end{aligned}
 \label{eq:score}
\end{equation}
All methods use the same assessment weights. $Q_\Theta$ requires every
nonzero-weight term to be available. Unlike $J$, it omits distance and gear
penalties and adds turning cost ($w_t=1$); it is a secondary diagnostic.
We separately measure route fractions violating mapped safe
clearance, observed pitch/roll allowances or $u_i=1$. An independent check
samples true arcs more finely, at no more than 1.25\,cm or $0.5^\circ$,
including endpoints.

For a bounded component $X$ with coverage $C_X$, full-route cost lies in
$[C_X\bar X,C_X\bar X+1-C_X]$ if observed estimates are held fixed.
We average these bounds per trial; classification error remains outside them.
All methods use a 1\,m goal tolerance. Failed and partial queries remain in
the attainment denominator. Length and cost means give equal weight to the
same goal-reaching trials for every method. Multiple configurations of one
map are not independent terrain sites.

Table~\ref{tab:methods} separates finite costs under shared hard constraints,
motion and search budgets. \emph{Center surface} isolates contact sampling;
\emph{Heading average} and \emph{Slope magnitude} test attitude preference.
\emph{Center-cell cost} also uses slope magnitude, but flat surface-test maps
isolate contact placement. Baseline prices distance; No surface cost omits
surface and uncertainty terms.

\begin{table}[t]
\centering\footnotesize
\setlength{\tabcolsep}{3pt}
\caption{Compared finite costs; hard body/support checks are shared.}
\label{tab:methods}
\begin{tabular}{@{}llll@{}}
\toprule
Method & Attitude & Body cost & Surface query \\
\midrule
Baseline & -- & -- & -- \\
No surface cost & Directional & -- & -- \\
Slope magnitude & Slope & -- & -- \\
Heading average & Mean & -- & Contacts* \\
Center-cell cost & Slope & -- & Center \\
Center surface & Directional & -- & Center \\
Body cost + center & Directional & Yes & Center \\
Body cost + contacts & Directional & Yes & Contacts \\
VCTP & Directional & -- & Contacts \\
\bottomrule
\end{tabular}
\par\smallskip
\parbox{\columnwidth}{Slope uses gradient magnitude; Mean averages all eight
attitude costs. Body cost denotes the additional soft body-based formulation,
not clearance checks. *Surface cost is active in the joint-policy bank,
but disabled in the cross-slope policy.}
\end{table}

\emph{Body cost + contacts} adapts RB-TRG's support, lateral-inclination and
interference costs~\cite{guo2026rbtrg}, averaging the mean and highest-cost
10\% of primitive samples equally. It receives VCTP's identical contact
surface term. \emph{Body cost + center} changes only surface sampling.
Local update and Full rebuild differ only in recomputation scope.

\subsection{Simulation and field-recording setup}
\label{sec:setup}

Table~\ref{tab:vehicles} declares the model and policies. All use
1\,m clearance, 0.20\,m support residual,
0.30\,m height departure, 2\,m/s nominal speed, $a_+=a_-=1.6$\,m/s$^2$
and $a_{\rm safe}=1$\,m/s$^2$. Larger virtual bodies or tighter attitude
allowances query the same observations more conservatively; deployment
remains limited by the actual Gator's capabilities.

\begin{table}[t]
\centering\footnotesize
\caption{Declared vehicle model and policy parameters.}
\label{tab:vehicles}
\begin{tabular}{@{}ll@{}}
\toprule
Body; mass & $2.4\times1.4$\,m; 700\,kg \\
Four contact offsets & $(\pm0.96,\pm0.60)$\,m; equal loads \\
Wheel radius; width & 0.275\,m; 0.20\,m \\
Turning radius & 3\,m (Ackermann) \\
\midrule
Policy: pitch / roll & Weights $(w_a,w_s,w_u,w_t)$ \\
\midrule
A: $20^\circ/15^\circ$ (sim.) & $(4,4,0.5,1)$ \\
B: $20^\circ/15^\circ$ (sim.) & $(4,0,0,1)$ \\
D: $20^\circ/15^\circ$ (recordings) & $(4,4,0,1)$ \\
E: $20^\circ/15^\circ$ (recordings) & $(4,0,0,1)$ \\
F: $5^\circ/15^\circ$ (recordings) & $(4,0,0,1)$ \\
\bottomrule
\end{tabular}
\end{table}

Each simulation family isolates a route-choice mechanism, with eight variants
on a 0.20\,m grid spanning
$40\times24$\,m. \emph{Surface avoidance} offers a short unfavorable strip and
a preferred bypass. \emph{Cross-slope routing} exchanges a direct off-camber
crossing for a longer bend. \emph{Contact placement} provides curved compacted
wheel strips on gravel and a paved bypass. These fully observed queries
share maps, endpoints, primitives and a 45\,s search budget, with two million
coarse and 750,000 vehicle-state expansions; field compilation is separate.

A joint-policy bank places grass between banked paved shoulders. Eight maps,
surface weights $\{0,4,16\}$, three pitch/roll allowances
($20^\circ/15^\circ$, $20^\circ/6^\circ$, $6^\circ/6^\circ$) and three
methods (Center surface, Heading average and VCTP) give 216 queries.
A second comparison fixes the surface weight at 4 and gives each formulation nine settings: VCTP attitude weights
$\{0.25,0.5,1,2,4,8,16,32,64\}$; body-cost weights $\{1,4,16\}$ crossed
with lateral priorities $\{1,2,4\}$. Body front/rear variation, inclination
and interference are normalized by body length times the tangent of safe
pitch, safe roll divided by lateral priority, and 0.30\,m, respectively,
then clipped at 3. Including its center-sampling control gives 648 queries.
All share a 15\,s budget and candidate-support checks at the actual headings.
We retain every setting rather than choosing a favorable one for each map.

\begin{figure}[t]
 \centering
 \includegraphics[width=0.78\columnwidth]{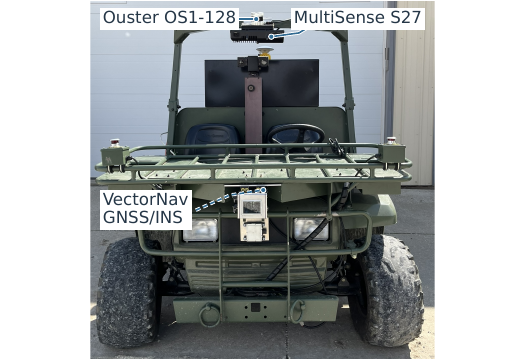}
 \caption{RGator sensing platform. The centrally housed GNSS/INS anchors
 initial gravity; LiDAR registration supplies the subsequent map trajectory.}
 \label{fig:hardware_setup}
\end{figure}

\begin{figure*}[t]
 \centering
 % Replace only the legacy title strip in the page layout; source pixels,
 % scene callouts and the image file remain unchanged. The strip is 44 px
 % at the source's 219.9894 dpi, i.e. 14.401 bp.
 {\sffamily\small
 \makebox[0.328\textwidth][l]{(a) Surface identification}\hfill
 \makebox[0.328\textwidth][l]{(b) Cross-slope route}\hfill
 \makebox[0.328\textwidth][l]{(c) Hidden object}\par}
 \smallskip
 \includegraphics[width=\textwidth,trim=0 0 0 14.401bp,clip]{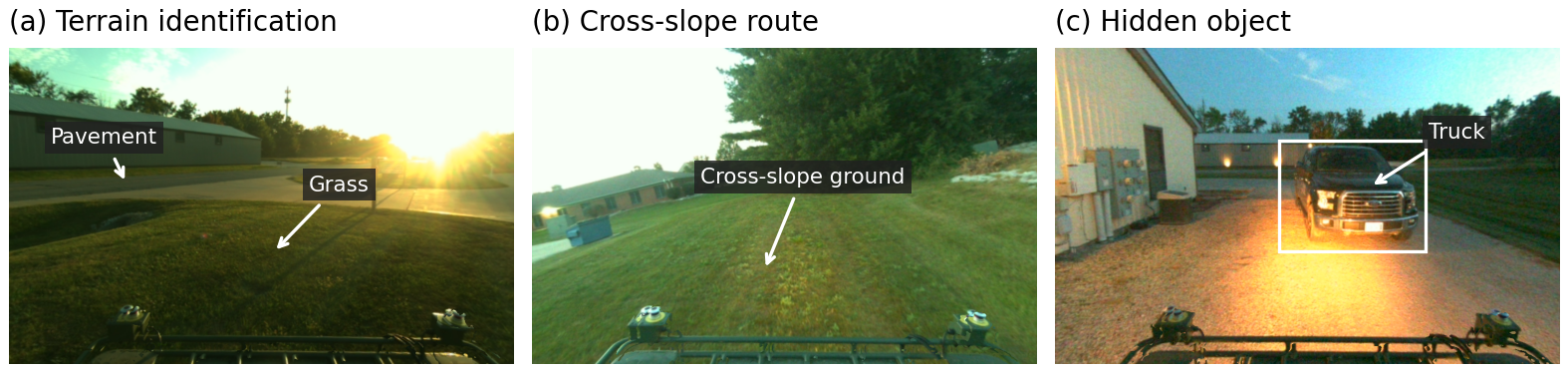}
 \caption{Recorded scenes: surface avoidance, cross-slope routing and
 a hidden object. Callouts identify scene features, not model predictions.
 Display-only exposure adjustment leaves classifier inputs unchanged.}
 \label{fig:field_scenes}
\end{figure*}

Offline planning uses 27 field-recorded RGator datasets from the platform in
Fig.~\ref{fig:hardware_setup} and scenes in
Fig.~\ref{fig:field_scenes}: ten surface
avoidance (five crossings, five bypasses), eleven cross-slope (five crossings,
six flatter bypasses) and six hidden-object recordings (five approaches,
one detour). Planners query the reconstructed maps; surface and cross-slope
queries use each recording's own endpoint.
Ouster OS1-128 scans at 10\,Hz
(1024 columns), MultiSense S27 RGB-D and VectorNav GNSS/INS provide the
recorded inputs. Processing uses an eight-core Xeon E-2278GE, RTX A6000
(48\,GB), Ubuntu 22.04 and ROS~2 Humble.

Each recording starts KISS-ICP 1.3.0~\cite{vizzo2023ral} with deskew and
0.5\,m registration voxels. Initial VectorNav attitude anchors gravity;
subsequent registration follows relative LiDAR motion. GOOSE-M2F Swin-Large
\cite{lingam2026goosem2f} runs on original RGB at $640\times384$, with flip
inference in FP32/TF32 at approximately one sampled frame per second.
Calibrated RGB-D projection and visibility-checked Ouster ground projection
associate the predicted IDs with 0.20\,m cells. The sampling rate describes
offline processing, not measured end-to-end live throughput.

Surface comparisons retain the archived maximum-height input, with only
missing cells in a gated 3\,m starting patch filled by a local plane.
Its inferred mask is excluded from coverage. Cross-slope and controlled
reveal queries use lower support with separate upper/lower spans and height
steps above 0.30\,m marking obstructions. Every paired method sees identical
inputs; assessment uses recorded upper and lower evidence.

Policy D uses $w_u=0$ to isolate surface/attitude preferences from
missing-observation penalties. Unfitted support receives only the distance multiplier, with
motion penalties and observed constraints retained. This cost comparison
does not authorize execution on unknown ground.
To assess sensitivity to that choice, a 400-query study crosses
$w_u\in\{0,0.25,0.5,1,2\}$ with all
four methods, both with the archived patch and with those inferred cells
restored to unknown. All four methods receive the same uncertainty term;
the main table retains its original policy D rather than substituting a
favorable sensitivity setting.

For hidden objects, we first compare selective and full evaluation as the
recorded maps grow. Their 164 prefixes give 158 consecutive updates per
configuration: VCTP under E and F, and Heading average under E. A separate
controlled reveal withholds and restores 417 measured truck-region cells in
the detour map, without adding ground. Three starting offsets,
six pitch limits (5--30$^\circ$) and three reveal positions give 54 cases.
At a fixed reveal plane, retained and fresh graphs receive the same current
pose, goal and changed observations. This isolates repair consistency, not
physical braking. A width sweep (1.4--3.0\,m) additionally queries the same
map under five body envelopes, six pitch limits and two reveal states.

\section{Results}
\label{sec:results}

Simulation isolates heading/contact mechanisms; field-recorded maps test
their planning consequences. Tables~\ref{tab:results_simulation}
and~\ref{tab:results_hardware} follow Sec.~\ref{sec:evaluation}:
$n$ counts all trials; Goal gives attainment. Length $L$, family cost and
$Q_\Theta$ are averaged over the same goal-reaching trials for every method.
Cost is $\bar T_{\rm known}$ for surface tests and $\bar F$ for cross-slope
routing, both in $[0,1]$. Coverage $C_S/C_Z$ reports the portions with
identified surfaces and observed geometry. Component costs, distance,
attainment and coverage are primary; $Q_\Theta$ is a secondary diagnostic.

\subsection{Simulation}

% Numerical rows preserved exactly from the long draft.
\begin{table}[t]
\centering\small
\setlength{\tabcolsep}{3pt}
\caption{Simulation policy comparisons.}
\label{tab:results_simulation}
\begin{tabular}{@{}lrrrrr@{}}
\toprule
Method & $n$ & Goal (\%) & $L$ (m) & Cost & $Q_\Theta$ \\
\midrule
\multicolumn{6}{@{}l}{\textit{Surface avoidance (A): $\bar T_{\rm known}$}} \\
Baseline & 8 & 100.0 & 31.77 & 0.173 & 0.1100 \\
Center-cell / VCTP* & 8 & 100.0 & 32.35 & 0.124 & 0.0888 \\
\midrule
\multicolumn{6}{@{}l}{\textit{Cross-slope route (B): $\bar F$}} \\
\shortstack[l]{Baseline /\\Heading average*} & 8 & 100.0 & 30.00 & 0.637 & 0.5093 \\
VCTP & 8 & 100.0 & 44.31 & 0.434 & 0.4202 \\
\midrule
\multicolumn{6}{@{}l}{\textit{Contact placement (A): $\bar T_{\rm known}$}} \\
Baseline & 8 & 100.0 & 30.00 & 0.137 & 0.0670 \\
Center-cell cost & 8 & 100.0 & 33.08 & 0.084 & 0.1189 \\
VCTP & 8 & 100.0 & 32.20 & 0.051 & 0.0808 \\
\bottomrule
\end{tabular}
\par\smallskip
\parbox{\columnwidth}{\footnotesize *Identical paths; $n$ is per method.
Surface and elevation coverage are 1.00 throughout.}
\end{table}

All main simulation queries reach their goals with complete coverage.
The two-track maps isolate contact placement.
Relative to Center-cell cost, VCTP lowers modeled surface cost by 39.3\% and route
length by 0.88\,m; seven maps improve and one worsens. Accumulated surface
cost also falls, from 2.764 to 1.634 cost-meters. The mean advantage remains
when any one map is omitted. Against Baseline, additional turning
raises $Q_\Theta$ from 0.0670 to 0.0808 despite lower surface exposure.

For cross-slope routing, VCTP reduces mean predicted attitude cost by 31.9\%
relative to Heading average and Slope magnitude, improving every map while
adding 14.31\,m. Both take the same direct paths as Baseline.
VCTP spreads lower attitude
demand over a longer route: accumulated cost is almost unchanged
(19.01 versus 19.10 cost-meters).

\begin{figure*}[t]
 \centering
 \includegraphics[width=\textwidth]{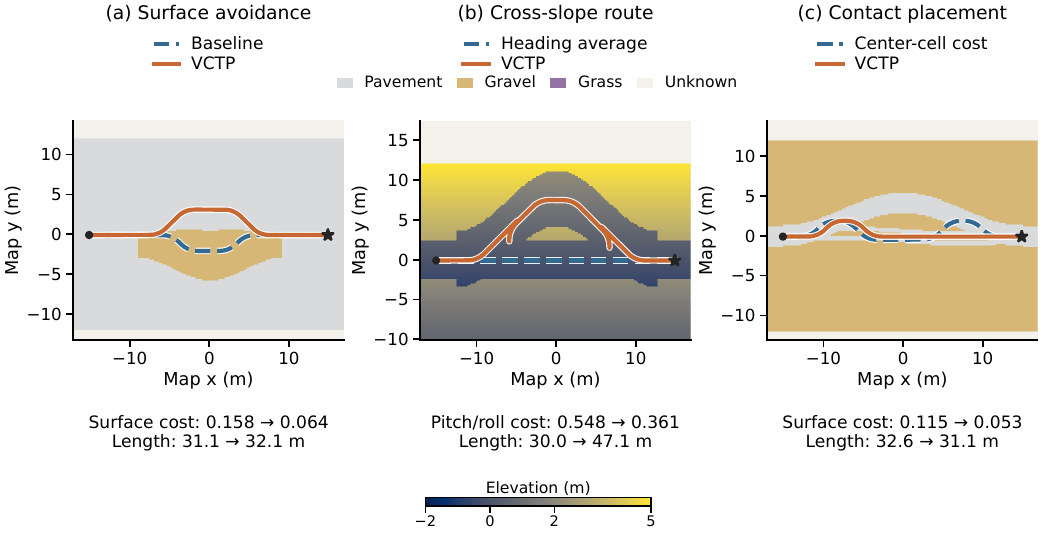}
 \caption{Simulation route comparisons: surface avoidance, cross-slope
 routing and contact placement. Panels show the first variant where all
 compared methods reach the goal and their paths differ. Circles and stars
 mark starts and goals. Table~\ref{tab:results_simulation} includes all variants.}
 \label{fig:simulation_results}
\end{figure*}

Broad surface-avoidance regions instead give identical Center-cell cost
and VCTP paths: contact placement adds no benefit here. Both reduce surface
cost by 28.3\% relative to Baseline for 0.58\,m of additional travel; four maps improve and four tie. No surface cost matches Baseline. This family tests
surface preference rather than contact placement.
Figure~\ref{fig:simulation_results} illustrates these mechanisms.

\noindent\textit{Joint policy and body-aware comparison.}
Combining surface and attitude preferences retains exact goal attainment
and complete coverage in all 216 joint-policy queries.
At surface weight 4 and $6^\circ/6^\circ$ attitude allowances, surface cost
is 0.0576 for VCTP, 0.0830 for Center surface and 0.0706 for Heading average;
lengths are 27.60, 24.00 and 25.46\,m. VCTP improves all eight maps against
center sampling and six against heading averaging, but turn cost is higher.
At nominal angles and weight 16, Heading average instead gives lower surface
cost (0.0393 versus 0.0474) but higher attitude cost (0.0685 versus 0.0465).

All 648 body-aware comparison queries also reach exact goals. Consider the
first declared body setting (weight 1, lateral priority 1) with tight angles:
Body cost + contacts gives surface cost 0.0591, attitude cost 0.00146 and
length 27.68\,m. VCTP at weight 4 gives 0.0576, 0.01294 and 27.60\,m.
Across the full set of settings, neither formulation dominates mean length,
surface cost and attitude cost. Contact sampling nevertheless benefits the
body-cost formulation too: mean surface cost falls at six of nine settings
under nominal limits and four of nine under each tighter policy; the rest
tie. At the tight-angle setting above, center sampling costs 0.0830 over
24\,m. Contacts improve seven maps and tie one, adding 3.68\,m on average.
The effect varies by map: at that setting with nominal limits, three improve
and five worsen despite the lower mean.

\subsection{Offline planning on field-recorded RGator data}

\begin{figure*}[t]
 \centering
 \includegraphics[width=\textwidth]{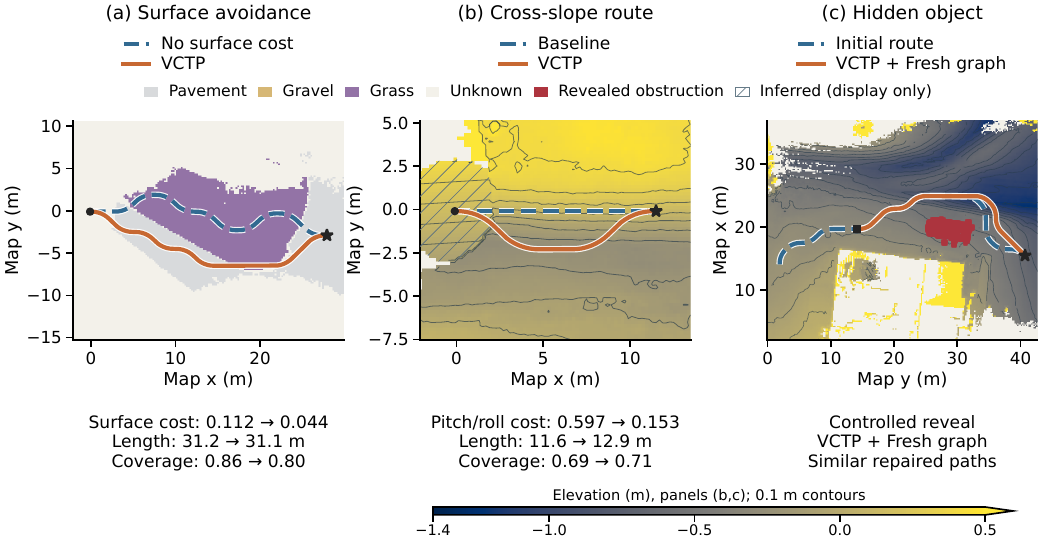}
 \caption{Offline routes on RGator recordings. (a) Surface avoidance on predicted IDs.
 (b) Cross-slope routing on lower support elevation. (c) Restoring measured
 truck-region cells (red) blocks the initial route's required body clearance.
 Retained and fresh graphs give the same repaired path from square to star.
 Panels (a,b) show the first recording where compared methods reach the goal
 on distinct paths; (c) shows the first nominal-policy controlled reveal.
 Elevation scales match; (c) swaps axes for display.}
 \label{fig:hardware_results}
\end{figure*}

Figure~\ref{fig:hardware_results} shows planned routes, distinct from the
recorded driving trajectories.

% Numerical rows preserved exactly from the long draft.
\begin{table}[t]
\centering\footnotesize
\setlength{\tabcolsep}{2.4pt}
\caption{RGator recordings: offline policy and map-update comparisons.}
\label{tab:results_hardware}
\begin{tabular}{@{}lrrrrrr@{}}
\toprule
Method & $n$ & Goal & $L$ & Cost & $Q_\Theta$ & $C_S/C_Z$ \\
 & & (\%) & (m) & & & \\
\midrule
\multicolumn{7}{@{}l}{\textit{Surface avoidance (D): $\bar T_{\rm known}$}} \\
Baseline & 10 & 90.0 & 38.59 & 0.063 & 0.0942 & 0.70 / 0.89 \\
No surface & 10 & 90.0 & 38.12 & 0.060 & 0.1002 & 0.77 / 0.90 \\
Center surface & 10 & 90.0 & 39.74 & 0.041 & 0.0937 & 0.58 / 0.90 \\
VCTP & 10 & 90.0 & 39.58 & 0.040 & 0.0945 & 0.58 / 0.89 \\
\midrule
\multicolumn{7}{@{}l}{\textit{Cross-slope route (E): $\bar F$}} \\
Baseline & 11 & 63.6 & 20.02 & 0.328 & 0.3310 & 0.64 / 0.81 \\
Slope mag. & 11 & 63.6 & 21.74 & 0.105 & 0.2107 & 0.61 / 0.80 \\
Heading avg. & 11 & 63.6 & 21.06 & 0.107 & 0.2047 & 0.59 / 0.80 \\
VCTP & 11 & 63.6 & 21.62 & 0.100 & 0.2067 & 0.62 / 0.81 \\
\midrule
\multicolumn{7}{@{}l}{\textit{Hidden object (E): original observation updates}} \\
Update & Rec. & Updates & Equal & Cells (\%) & Time (s) & \\
Full rebuild & 6 & 158 & 158/158 & 100.0 & 0.194 & \\
Local update & 6 & 158 & 158/158 & 33.9 & 0.118 & \\
\bottomrule
\end{tabular}
\par\smallskip
\parbox{\columnwidth}{No surface = No surface cost; Slope mag. = Slope
magnitude. Update times are median offline field compilation times;
updates are not additional trials.}
\end{table}

All methods reach nine of ten surface goals and seven of eleven cross-slope
goals. The failed surface query exhausts its budget after clearance rejection.
Cross-slope failures include two restricted starts and two partial routes
outside the goal tolerance. All count toward attainment.

\noindent\textit{Surface avoidance.}
VCTP’s mean modeled cost is close to that of Center surface (0.040 versus 0.041): four recordings improve, one ties and four regress.
Removing one recording can reverse the mean advantage, and $Q_\Theta$ is
slightly higher.
Relative to Baseline, VCTP lowers cost over identified contacts by 36.5\%,
adding 0.99\,m; coverage is 0.58 versus 0.70. It improves eight
of nine recordings against No surface cost, whose coverage is 0.77.
The corresponding full-route bounds overlap:
$[0.024,0.446]$ for VCTP and $[0.048,0.281]$ for No surface cost.
The identified-contact gain therefore does not establish a full-route gain.

The uncertainty robustness analysis increases coverage while retaining nine
goals at every setting.
At $w_u=0.5$, VCTP and Center surface have nearly equal
coverage (0.818/0.819), costs 0.0407/0.0423 and lengths 38.44/38.81\,m.
At this matched coverage, the small mean advantage remains when any one
recording is omitted. Restoring
the inferred starting patch to unknown gives the same ordering:
coverage 0.8155/0.8158 and costs 0.0413/0.0424. The table retains policy D;
these are exploratory settings. Stronger uncertainty penalties can erase
surface differences: at $w_u=2$, VCTP and No surface cost both give
approximately 0.0539. Missing-cost bounds overlap throughout.

\noindent\textit{Cross-slope routing.}
VCTP's predicted attitude cost on observed support is 4.8\% below Slope
magnitude at nearly equal length and 6.5\% below Heading average; its
$Q_\Theta$ is nevertheless 1.0\% higher than the latter's.
Against Baseline, cost falls from 0.328 to 0.100, improving all seven common goal-reaching trials for 1.60\,m of extra travel.
Elevation coverage is about 0.81. Full-route bounds overlap:
$[0.079,0.273]$ for VCTP and $[0.259,0.452]$ for Baseline.

The recorded driving trajectories also let us compare estimated roll with
VectorNav measurements. Lower-support estimates have mean absolute errors
of $0.85^\circ$ across five crossings and $0.92^\circ$ across six bypasses,
versus $1.14^\circ/1.53^\circ$ from maximum height. Eligible moving-distance
coverage is 0.75/0.87. The estimates share initial gravity alignment, and this
check concerns recorded trajectories, not the alternative planned paths.
The body-cost comparator requires height inputs unavailable in these
recorded-map queries, so its evaluation remains simulation-only.

\noindent\textit{Hidden object and vehicle configuration.}
Selective and full evaluation agree in all 474 comparisons on the original
map prefixes: restriction masks, reasons and eight-heading costs match.
These comparisons reuse six recordings across configurations, rather than
adding independent trials. Under E, selective evaluation recomputes 66.1\%
fewer cells. Median field
compilation takes 0.118 versus 0.194\,s; these times exclude the rest of the
pipeline. The separate controlled reveals give 36 cases with no initial
route, three retained routes and 15 repairs. All 15 repairs reach the exact
goal with identical paths from retained and fresh graphs and fully observed
support. Selective support evaluation preserves these outcomes. The
maximum-height representation yields no initial route in all 54 cases;
using lower support restores routes while retaining the measured obstruction.

Changing the vehicle width changes its route on the same observations.
For vehicle widths of 1.4--2.6\,m, planned routes reach the goal under all six pitch policies before and after the reveal. At nominal pitch, post-reveal length grows from 33.38 to
37.17\,m between 1.4 and 2.6\,m widths. The 3.0\,m envelope rejects the
fixed start in all twelve queries. This is a start-clearance result, not
proof that every route is blocked for that vehicle.

\noindent\textit{Available-geometry and motion checks.}
All 160 returned candidates from the 172 main queries pass the assessed
clearance, attitude, turning-radius and lateral-acceleration checks over
437,272 fine arc samples plus endpoints. The joint-policy, body-aware,
uncertainty and controlled-repair studies likewise have no assessed
violations. These checks use available map evidence and cannot verify
missing support or perception accuracy.

\section{Discussion}

The fully observed two-track experiment isolates the benefit of contact
placement: it lowers modeled surface cost relative to center sampling and
shortens the route. Broad surface regions instead test surface preference,
while cross-slope maps isolate heading dependence.
Contact sampling also lowers mean modeled surface cost within the adapted
body-cost comparator at several tested settings, indicating that the benefit
can extend beyond VCTP's particular attitude formulation. The effect varies
between maps and settings; neither formulation dominates all reported tradeoffs.

Field-recorded maps show that declared preferences change routes and lower
modeled costs over observed portions relative to Baseline, with extra travel
and unchanged attainment. Contact sampling has mixed field outcomes.
%Matched-coverage sensitivity is supportive but
%exploratory, not a replacement primary result.
Although the matched-coverage sensitivity analysis supports our findings, it should be viewed as an exploratory rather than a primary result.

Incomplete coverage leaves full-route rankings unresolved, and alternative planned routes were not executed. The physical response to the soil and closed-loop execution require separate validation.

\section{Conclusion}

VCTP evaluates wheel placement and predicted support attitude at each heading
while enforcing body-clearance limits. Offline comparisons on
RGator recordings show lower modeled costs over observed portions than
distance-focused planning, with additional travel and unchanged goal
attainment; full-route rankings remain unresolved. On the tested two-track
maps, evaluating surfaces at wheel contacts improves the modeled
surface-cost/length tradeoff relative to center sampling. Selective evaluation
matches full-field recomputation and preserves the tested controlled-repair
outcomes.

\section*{Acknowledgment}

This work was supported by the U.S. Army Corps of Engineers,
Engineer Research and Development Center, Construction
Engineering Research Laboratory, under Contract Award
No.~W9132T249C011.

\balance
\bibliographystyle{IEEEtran}
\bibliography{references}

\end{document}